# A DH-parameter based condition for 3R orthogonal manipulators to have 4 distinct inverse kinematic solutions

P. Wenger, D. Chablat and M. Baili

Institut de Recherche en Communications et Cybernétique de Nantes UMR CNRS 6597

1, rue de la Noë, BP 92101, 44312 Nantes Cedex 03 France

*Positioning 3R manipulators may have two or four inverse kinematic solutions (IKS). This paper derives a necessary and sufficient condition for 3R positioning manipulators with orthogonal joint axes to have four distinct IKS. We show that the transition between manipulators with 2 and 4 IKS is defined by the set of manipulators with a quadruple root of their inverse kinematics. The resulting condition is explicit and states that the last link length of the manipulator must be greater than a quantity that depends on three of its remaining DH-parameters. This result is of interest for the design of new manipulators.*

## 1 Introduction

This paper focuses on positioning 3R orthogonal manipulators i.e. positioning 3R manipulators with orthogonal joint axes. A positioning manipulator may be used as such for positioning tasks in the Cartesian space (x, y, z), or as a regional structure of a 6R manipulator with spherical wrist. Among the various kinematic criteria that can be used to assess the performances of a manipulator, the accessibility inside the workspace, i.e. the number of inverse kinematic solutions (IKS) in the workspace, is of primary interest. Positioning 3R manipulators are known to have at most four inverse kinematic solutions (IKS) in their workspace [1]. In general, the number of IKS varies from one point to another in the workspace [2-5], which may include regions with 0, 2 or 4 IKS [6-8]. Depending on its geometric parameters, a 3R manipulator may be *binary*, i.e. may have at most two IKS in its workspace, or it may be *quaternary*, i.e. it may have up to four IKS [1]. We know from [9] that 3R manipulators with any two intersecting joint axes (i.e. $a_1=0$ or $a_2=0$ or $a_3=0$) are quaternary; [13] showed that a 3R orthogonal manipulator with no offset at joint 3 (i.e. $d_3=0$) is quaternary if the last link length is greater than the second one (i.e. $a_3>a_2$). But this condition is not necessary, that is, a manipulator such that $d_3=0$ may be quaternary even if $a_3<a_2$. On the other hand, [10] stated a particular necessary and sufficient condition, namely, 3R orthogonal manipulators with no joint offsets (i.e. $d_2=d_3=0$) are quaternary if, and only if, $a_1≠a_2$ and the link lengths do not satisfy $a_1>a_2>a_3$. To the authors' knowledge, no more general DH-parameter based necessary and sufficient condition has been derived for a manipulator to be quaternary.

This paper derives an explicit DH-parameters based necessary and sufficient condition for a 3R manipulator with orthogonal joint axes to be quaternary. We show that the transition between binary and quaternary manipulators is defined by the set of manipulators with a quadruple root of their inverse kinematics. The set of such manipulators is shown to form a separating surface in the manipulator parameter space, which can be defined explicitly by an equation of the form $a_3=f(a_1, a_2, d_2)$. This result is of interest for the design of new manipulators.

## 2. Preliminaries

### 2.1 Manipulators under study

The length DH-parameters of an orthogonal manipulator are referred to as $a_1$, $a_2$, $a_3$, $d_1$, $d_2$, $d_3$ while the angle parameters $α_1$ and $α_2$ are set to –90° and 90°, respectively. From now on, $a_1$ can be set to 1 without loss of generality. First joint offset $d_1$ can be chosen equal to zero by an appropriate choice of the reference frame. Last joint offset $d_3$ is set to 0 in this study. Thus, we need to handle only three design parameters, which will assume strictly positive values in this study. Fig. 1 shows the kinematic architecture of an orthogonal manipulator in its zero configuration. The three joint variables are referred to as $\theta_1$, $\theta_2$ and $\theta_3$, respectively. The position of the end-tip (or wrist center) is defined by the three Cartesian coordinates $x$, $y$ and $z$ of the operation point $P$ with respect to a reference frame (O, $\mathbf{x}$, $\mathbf{y}$, $\mathbf{z}$) attached to the manipulator base as shown in Fig.1.

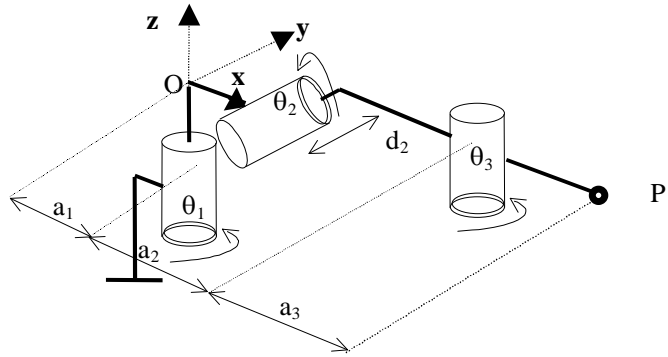

**Fig. 1 : Orthogonal manipulator in its zero configuration**

## 2.2 Singularities curves in the joint space and in the workspace

The singularities of general 3R manipulators have been derived in [10,12]. They can be determined by calculating the determinant of the Jacobian matrix [10], or using a recursive method [12]. For the orthogonal manipulators under study, i.e. with $\alpha_1$, $\alpha_2$, $a_1$ and $d_3$ equal to –90°, 90°, 1 and 0, the determinant of the Jacobian matrix takes the following form [10]:

$$\det(\mathbf{J}) = (a_2 + c_3 a_3)(c_2(s_3 a_2 - c_3 d_2) + s_3 a_1)$$

where $c_i = \cos(\theta_i)$ and $s_i = \sin(\theta_i)$. A singularity occurs when $\det(\mathbf{J})=0$. Since the singularities are independent of $\theta_1$, the contour plot of $\det(\mathbf{J})=0$ can be displayed as curves in $-\pi \leq \theta_2 < \pi, -\pi \leq \theta_3 < \pi$. The singularities can also be displayed in the Cartesian space by plotting the points where the inverse kinematics has double roots [3,7]. Thanks to their symmetry about the first joint axis, it is sufficient to draw a half cross-section of the workspace by plotting the points ( $\rho = \sqrt{x^2 + y^2}$ , $z$).

Two cases arise:

- if $a_2 > a_3$, the first factor of $\det(\mathbf{J})$ cannot vanish and the singularities form two distinct curves $S_1$ and $S_2$ in the joint space [10]. When the manipulator is in such a singularity, there is line that passes through the operation point and that cuts all joint axes [12 ]. The singularities form two disjoint sets of curves in the workspace. These two sets define the internal boundary $WS_1$ and the external boundary $WS_2$, respectively, with $WS_1=f(S_1)$ and $WS_2=f(S_2)$. Figure 2($a$) shows the singularity curves when $a_2=2$, $a_3=1.5$, $d_2=1$. For this manipulator, the internal boundary $WS_1$ has four cusp points, where three IKS coincide [7]. It divides the workspace into one region with two IKS (the outer region) and one region with four IKS (the inner region), which means that this manipulator is quaternary. As shown in section 3, the left and right segments of the internal boundary may cross and define a void when $d_2$ is decreased; if $d_2$ is sufficiently small, the internal boundary has no cusp, the region with four IKS is replaced with a void and the manipulator is binary.

- if $a_2 \leq a_3$, the operation point can meet the second joint axis whenever $\theta_3 = \pm\arccos(-a_2/a_3)$ and two horizontal

lines appear in the joint space. No additional curve appears in the workspace cross-section but only two points. This is because, since the operation point meets the second joint axis when $\theta_3 = \pm \arccos(-a_2/a_3)$, the location of the operation point does not change when $\theta_2$ is rotated. Figure 2(b) shows the singularity curves of a manipulator such that $a_2 = 3$, $a_3 = 4$, $d_2 = 3$.

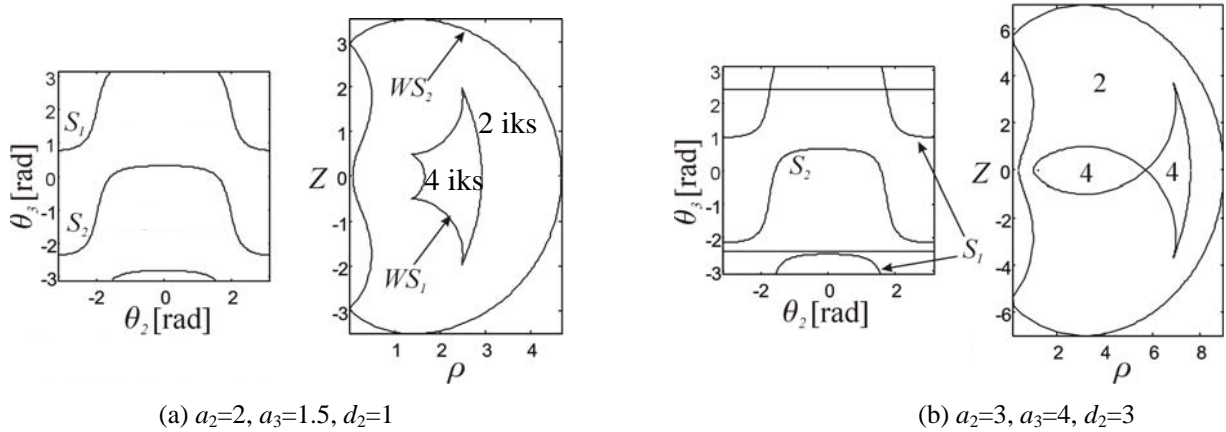

(a) $a_2 = 2$, $a_3 = 1.5$, $d_2 = 1$                                    (b) $a_2 = 3$, $a_3 = 4$, $d_2 = 3$

**Fig. 2 : Singularity curves for a quaternary manipulator when $a_2 > a_3$ (a) and when $a_2 < a_3$ (b)**

## 3. Transition between binary and quaternary manipulators

In this section, we show that the transition between binary and quaternary manipulators is the set of manipulators with a quadruple root of their inverse kinematics. This result is a consequence of a classification work conducted in [13,14]. Using Groebner Bases and Cylindrical Algebraic Decomposition, [14] derived the equations of several surfaces that divide the DH-parameters space into 105 domains of manipulators having the same number of cusps in their workspace. The systematic investigation of the 105 domains and their kinematic interpretation conducted in [13] showed that, (*i*) all manipulators satisfying $a_2 \leq a_3$ are quaternary, and (*ii*) the set of manipulators satisfying $a_2 > a_3$ is composed of two adjacent domains, one of which being the set of all binary manipulators, the other one being composed of only quaternary manipulators with four cusps like the one shown in Fig. 2(*a*). In other words, binary manipulators exist only when $a_2 > a_3$ and a boundary surface exists that divides the set of manipulators such that $a_2 > a_3$ into two domains in the parameter space ($a_2$, $a_3$, $d_2$). Now, we show in Fig. 3 how a quaternary manipulator with four cusps turns binary under the continuous deformation of the internal boundary of its workspace as $a_3$ is progressively decreased ($a_2 = 1.5$, $d_2 = 0.5$). In Fig. 3(*a*), $a_3 = 1.1$, the manipulator is quaternary and the internal boundary is like in Fig. 2(*a*) with four cusps and no void. When $a_3 = 0.9$, the two lateral segments cross. Two nodes appear, which define a void and two separate regions with four IKS (Fig. 3(*b*)). The two cusps and the node of each such region get closer to each other as $a_3$ is decreased (Figs. 3(*c*) and 3(*d*)). Then, they merge into one unique point with four coincident IKS (the region with four IKS is reduced to one point) and, finally, disappear and the manipulator turns binary (Fig 3(*e*)). Thus, the transition between a quaternary manipulator and a binary manipulator is characterized by the existence of a pair of four coincident IKS.

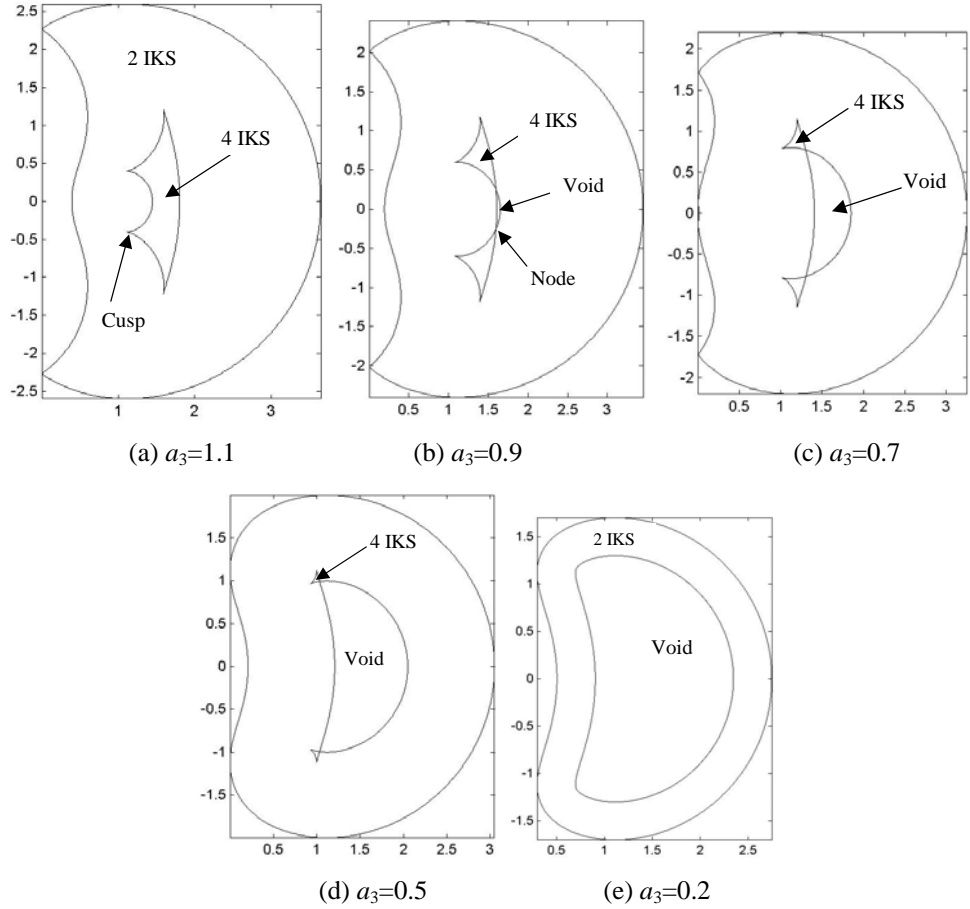

Fig. 3 : **Continuous deformation of the internal boundary as $a_3$ is decreased ($a_2$=1.5, $d_2$=0.5). From 1.1 to 0.5, the manipulator is quaternary (a-c). From 0.5 to 0.2, two cusps and one node merge into one point with four equal IKS and then disappear : the manipulator turns binary (d-e).**

## 4. Existence condition of a point with four IKS

To get the equation of the separating surface, we derive the existence condition of a point with four IKS. This can be done with the fourth-degree inverse kinematics univariate polynomial in $t$=tan($\theta_3$/2). The fourth-degree inverse kinematics polynomial of 3R manipulators was derived in [3]. It can be set in the form $P(t) = C_0 t^4 + 4C_1 t^3 + 6C_2 t^2 + 4C_3 t + C_4 = 0$ where $C_0$, $C_1$, $C_2$ and $C_4$ are functions of $\rho = \sqrt{x^2 + y^2}$, $z$ and the DH-parameters [3]. For the orthogonal manipulators under study, i.e. $\alpha_1$, $\alpha_2$, $a_1$ and $d_3$ equal to –90°, 90°, 1 and 0, respectively, $C_0$, $C_1$, $C_2$ and $C_4$ can be written as:

$$C_0 = a_2^2 \, a_3^2 - a_2 \, a_3 \, V - R + \frac{V^2}{4} + d_2^2$$

$$C_1 = \frac{a_3 d_2 \, (-2 a_2 a_3 + V + 2)}{2}$$

$$C_2 = -\frac{a_2^2 \, a_3^2}{3} + \frac{2 a_3^2 \, d_2^2}{3} + \frac{2 a_3^2}{3} - \frac{R}{3} + \frac{V^2}{12} + \frac{d_2^2}{3}$$

$$C_3 = \frac{a_3 d_2 \, (2 a_2 a_3 + V + 2)}{2}$$

$$C_4 = a_2^2 \, a_3^2 + a_2 \, a_3 \, V - R + \frac{V^2}{4} + d_2^2$$

Wenger P., Chablat D. et Baili M., "A DH-parameter based condition for 3R orthogonal manipulators to have 4 distinct inverse kinematic solutions", Journal of Mechanical Design, Volume 127, pp. 150-155, Janvier 2005.

where $V = -x^2 - y^2 - z^2 - 1 + a_2^2 + d_2^2 + a_3^2$ and $R = x^2 + y^2$.

[3] also derived the existence conditions of multiple IKS. For $P(t)$ to have four equal roots, the following three equations must be simultaneously satisfied [3]:

$$C_0 C_4 - 4 C_1 C_3 + 3 C_2^2 = 0 \qquad (1)$$

$$C_0 C_2 C_4 + 2 C_1 C_2 C_3 - C_0 C_3^2 - C_1^2 C_4 - C_2^3 = 0 \qquad (2)$$

$$C_0 C_2 - C_1^2 = 0 \qquad (3)$$

We need to eliminate the Cartesian coordinates $x$, $y$ and $z$ in order to write a condition on the DH-parameters only. Thus, $V$ and $R$ must be eliminated. This task is performed using computer algebra tools [11]. Such tools are available in symbolic commercial softwares. We have used the Maple function *resultant*. First, $R$ is eliminated from (1) and (3). This yields a fourth-degree polynomial in $V$. Then, $R$ is eliminated from (2) and (3). We get a third-degree polynomial in $V$. Finally, $V$ is eliminated from the aforementioned two polynomials. The resulting polynomial is:

$$a_2{}^{12} a_3{}^4 d_2{}^4 Q_1 Q_2 Q_3 \qquad (4)$$

where $Q_1$, $Q_2$ and $Q_3$ are polynomials in $a_2$, $a_3$ and $d_2$:

$$
\begin{aligned}
Q_1 = &\, a_2{}^6 a_3{}^2 - 2 a_2{}^4 a_3{}^2 - a_2{}^4 a_3{}^4 + 3 d_2{}^2 a_2{}^4 a_3{}^2 + a_2{}^2 a_3{}^2 + 2 a_2{}^2 a_3{}^4 \\
&- 2 d_2{}^2 a_2{}^2 a_3{}^4 + 3 d_2{}^4 a_2{}^2 a_3{}^2 - a_3{}^4 + d_2{}^2 a_3{}^2 - 2 d_2{}^2 a_3{}^4 + 2 d_2{}^4 a_3{}^2 - d_2{}^4 a_3{}^4 + d_2{}^6 a_3{}^2 - d_2{}^2 a_2{}^2
\end{aligned}
\qquad (5)
$$

$$
\begin{aligned}
Q_2 = &\, -543 a_3{}^5 a_2{}^2 d_2{}^2 + 648 a_3{}^5 a_2{}^4 - 81 a_3{}^5 a_2{}^2 - 32 a_3{}^4 d_2{}^2 a_2 - 8 a_3{}^4 d_2{}^2 a_2 + 1110 a_3{}^3 a_3{}^4 d_2{}^2 - 25 d_2{}^2 a_2{}^5 \\
&- 47 a_2{}^2 a_3{}^3 d_2{}^2 - 141 a_2{}^2 d_2{}^4 a_3{}^3 - 47 a_2{}^2 a_3{}^3 d_2{}^8 - 210 a_3 a_3 d_2{}^2 a_2{}^6 + 486 a_3{}^3 a_2{}^{10} + 972 a_2{}^6 a_3{}^3 \\
&- 1215 a_3{}^3 a_2{}^8 - 1458 a_2{}^5 a_3{}^4 - 8 a_2 a_2 d_2{}^{10} a_3{}^4 - 48 a_2 a_3{}^4 d_2{}^6 - 32 a_2 a_3{}^4 d_2{}^8 - 141 a_2{}^7 d_2{}^2 a_3{}^3 \\
&+ 243 a_3{}^4 a_2{}^3 + 81 a_3{}^2 a_2{}^5 - 162 a_3{}^2 a_2{}^7 + 81 a_3{}^3 a_2{}^9 - 243 a_3{}^3 a_2{}^4 + 1224 a_3{}^4 d_2{}^6 d_2{}^3 + 300 a_3{}^4 d_2{}^8 a_3 \\
&+ 1791 a_3{}^4 d_2{}^4 a_2{}^3 - 801 a_2{}^4 a_3{}^3 d_2{}^2 + 35 a_2{}^2 a_3 d_2{}^4 + 29 d_2{}^6 a_2{}^3 a_3{}^3 + 444 a_3{}^2 d_2{}^2 a_2{}^5 + 58 a_3{}^2 d_2{}^4 a_2{}^3 \\
&+ 35 a_3 d_2{}^2 a_2{}^4 + 16 d_2{}^{12} a_3{}^5 + 80 a_3{}^5 d_2{}^{10} + 160 a_3 d_2{}^5 + 160 d_2{}^6 a_3{}^5 + 80 a_3{}^5 d_2{}^4 + 16 a_3{}^5 d_2{}^2 \\
&- 177 a_3{}^3 d_2{}^4 a_2{}^5 - 813 a_3{}^3 d_2{}^2 a_2{}^7 - 2340 a_3{}^8 a_3{}^5 d_2{}^2 + 2052 a_3{}^5 d_2{}^2 a_2{}^4 - 2025 a_3{}^6 a_3{}^5 + 3078 a_3{}^8 a_3{}^5 \\
&+ 1872 d_2{}^2 a_2{}^8 a_3{}^5 - 3096 d_2{}^3 a_2{}^7 a_3{}^4 + 1440 d_2{}^6 a_2{}^5 a_3{}^4 - 459 a_3{}^3 a_2{}^9 d_2{}^4 + 72 a_3{}^4 a_2{}^9 d_2{}^2 \\
&+ 2268 a_2{}^{10} a_3{}^5 d_2{}^2 + 648 a_2{}^{12} a_3{}^5 + 972 a_2{}^{11} a_3{}^4 - 2268 a_2{}^{10} a_3{}^5 + 3159 a_2{}^7 a_3{}^4 - 1188 a_2{}^5 a_3{}^4 d_2{}^4 \\
&+ 260 1 a_3{}^6 a_3{}^3 d_2{}^2 - 72 d_2{}^6 a_2{}^5 a_3{}^5 + 180 a_3{}^3 d_2{}^6 a_2{}^3 + 2340 a_3{}^5 d_2{}^4 a_2{}^6 + 2352 a_3{}^5 d_2{}^2 a_2{}^4 \\
&- 1773 a_3{}^3 d_2{}^2 a_2{}^8 - 2682 a_2{}^5 a_3{}^4 d_2{}^2 - 1557 a_3{}^3 d_2{}^2 a_2{}^6 + 1872 a_2{}^7 a_3{}^4 d_2{}^2 - 2916 a_3{}^9 a_2{}^4 \\
&+ 29 a_2{}^3 a_3{}^3 d_2{}^2 - 168 a_2{}^3 a_3{}^3 d_2{}^8 - 552 a_2{}^2 d_2{}^8 a_3{}^3 - 1227 a_3{}^3 d_2{}^4 a_2{}^2 - 1233 a_2{}^2 a_3{}^5 d_2{}^6 \\
&- 84 a_3{}^5 d_2{}^{10} a_2{}^2 - 1845 a_3{}^3 d_2{}^4 a_2{}^4 - 1287 d_2{}^6 a_2{}^4 a_3{}^3
\end{aligned}
\qquad (6)
$$

$$Q_3 = -a_2 + a_3 d_2{}^2 + a_3 \qquad (7)$$

Note : we would have obtained exactly the same equations for $d_3 \neq 0$. In effect, it turns out that, when $d_3 \neq 0$, coefficients $C_0$, $C_1$, $C_2$ and $C_4$ have the same expressions as function of $V$ and $R$. This is because $d_3$ appears only in $V$ (more precisely, $V(d_3 \neq 0) = V(d_3 = 0) + d_3^2$). Since $V$ eliminated in Eqs. (1-3), the resulting condition does not change. Thus, the condition for a manipulator to have four equal IKS is independent of $d_3$.

## 5. Separating surface and the necessary and sufficient condition

Since elimination may generate spurious solutions [11], solutions of (4) include, in addition to the surface that separates quaternary and binary manipulators in the parameter space ($a_2$, $a_3$, $d_2$), other non-separating surfaces. We know that the surface that separates quaternary and binary manipulators is necessarily among the surfaces found in [14]. This is because [14] determined the surfaces that divide the parameter space into domains where the number of cusps is

Wenger P., Chablat D. et Baili M., "A DH-parameter based condition for 3R orthogonal manipulators to have 4 distinct inverse kinematic solutions", Journal of Mechanical Design, Volume 127, pp. 150-155, Janvier 2005.

constant. The equations of these surfaces are [14]:

$$a_2^2 - a_3^2 + d_2^2 = 0 \tag{8}$$

$$a_3^2 a_2^6 - a_3^4 a_2^4 + 3a_3^2 a_2^4 d_2^2 - 2a_3^3 a_2^4 + 2a_3^4 a_2^2 - 2a_3^4 a_2^2 d_2^2 + a_3^2 a_2^2 + 3a_3^2 a_2^2 d_2^4 - a_2^2 d_2^2 - 2a_3^4 d_2^2$$
$$-a_3^4 + a_3^2 d_2^6 + a_3^2 d_2^2 + 2a_3^2 d_2^4 = 0 \tag{9}$$

$$a_2^2 d_2^2 + a_2^2 - 2a_2^3 + a_2^4 - a_3^2 + 2a_2 a_3^2 - a_2^2 a_3^2 = 0 \tag{10}$$

$$a_2^2 d_2^2 + a_2^2 + 2a_2^3 + a_2^4 - a_3^2 + 2a_2 a_3^2 - a_2^2 a_3^2 = 0 \tag{11}$$

Comparing Eqs. (8-11) with Eqs. (4-7) show that Eq. (5), i.e. $Q_1=0$, is the same as Eq. (9). On the other hand, Eqs. (6) and (7) are different from Eqs. (8), (10) and (11). Thus, the only valid solution is $Q_1=0$. This equation can be put in an explicit form. In effect, this is a second-degree polynomial in $a_3^2$. Solving this quadratics for $a_3$ yields the equations of two regular surfaces given by the following two explicit equations:

$$a_3 = \frac{1}{2}\sqrt{2a_2^2 + 2d_2^2 - \frac{2((a_2^2 + d_2^2)^2 - (a_2^2 - d_2^2))}{AB}} \tag{12}$$

and

$$a_3 = \frac{1}{2}\sqrt{2a_2^2 + 2d_2^2 + \frac{2((a_2^2 + d_2^2)^2 - (a_2^2 - d_2^2))}{AB}} \tag{13}$$

where

$$A = \sqrt{(a_2+1)^2 + d_2^2} \quad \text{and} \quad B = \sqrt{(a_2-1)^2 + d_2^2}$$

Fig. 4 shows the graph of the aforementioned two surfaces. For more clarity, 2-dimensional sections of the surfaces are drawn in ($a_2$, $a_3$) for $d_2=0.5$ and $d_2=1$, respectively (Figs. 444444($a$) and 4($b$)), and in ($d_2$, $a_3$) for $a_2=0.5$ and $a_2=1.5$, respectively (Figs. 4($c$) and 4($d$)).

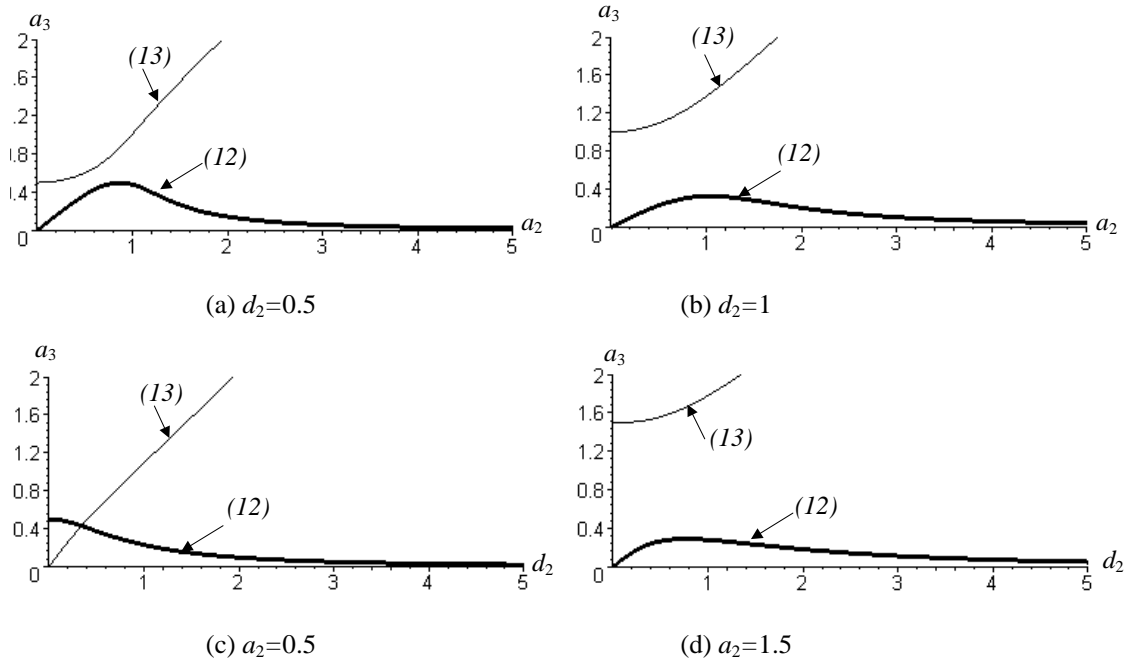

(a) $d_2=0.5$

(b) $d_2=1$

(c) $a_2=0.5$

(d) $a_2=1.5$

**Fig. 4 : Graphs of Eqs. (12) and (13) shown in sections of the DH-parameter space. Graph of (12) is shown in bold lines**

In sections shown in Figs. 4(a), 4(b) and 4(d), the boundary between quaternary and binary manipulators is defined only by Eq. (12) since Eq. (13) has solutions only for $a_3 > a_2$, i.e. for quaternary manipulators. In section $a_2$=0.5, on the other hand (Fig. 4(c)), the two graphs intersect and (13) has solutions in $a_3 < a_2$ when $d_2$ is small enough. It turns out, however, that the surface defined by Eq. (13) does not play any role in the separation and the really separating surface is defined by Eq. (12) only. In effect, let choose three test manipulators (1), (2) and (3) in section $a_2$=0.5, defined by ($a_3$=0.15, $d_2$=0.21), ($a_3$=0.4, $d_2$=0.1) and ($a_3$=0.45, $d_2$=0.4), respectively (Fig. 5a). These manipulators were chosen such that (1) and (2) are separated by (13), and (2) and (3) are separated by (12). Figure 5b shows the workspace of the three test manipulators. Manipulators (1) and (2) are binary whereas (3) is quaternary. Thus, the boundary surface that separates the binary from the quaternary manipulators is defined by (12) and the necessary and sufficient condition for an orthogonal manipulator to be quaternary is $a_3 > \dfrac{1}{2}\sqrt{2a_2{}^2 + 2d_2{}^2 - \dfrac{2((a_2{}^2 + d_2{}^2)^2 - (a_2{}^2 - d_2{}^2))}{AB}}$ .

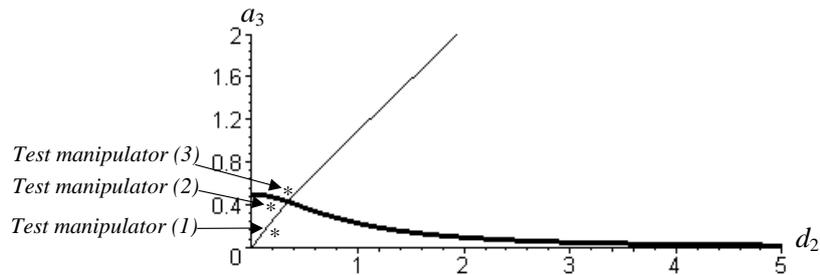

(a) Test manipulators in section $a_2$=0.5

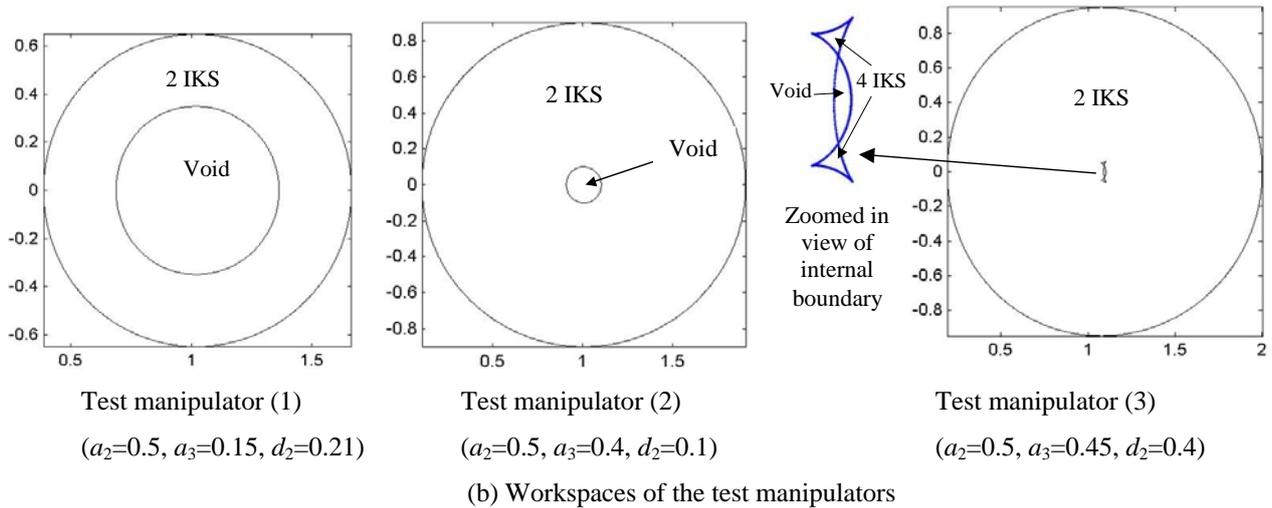

(b) Workspaces of the test manipulators

**Fig. 5 : The three test manipulators (a) and their workspace (b). Manipulators (1) and (2) are binary whereas (3) is quaternary**

For verification purposes, we have written a procedure that plots all binary manipulators by scanning the parameter space. The procedure checks the existence of a cusp point in a cross section of the workspace: it scans the internal boundary and checks the existence of a triple root of the inverse kinematics polynomial [3]. If there is no cusp and if $a_3 < a_2$, the manipulator is binary and a mark is plotted. As soon as a cusp is found, no mark is plotted and next

Wenger P., Chablat D. et Baili M., "A DH-parameter based condition for 3R orthogonal manipulators to have 4 distinct inverse kinematic solutions", Journal of Mechanical Design, Volume 127, pp. 150-155, Janvier 2005.

manipulator is checked. Figure 6 depicts the resulting plots in the same sections as in Fig. 4. Each parameter was scanned with a step of 0.03. Comparison of Figs. 6($a$), 6($b$), 6($c$) and 6($d$) with Figs. 4($a$), 4($b$), 4($c$) and 4($d$), respectively, confirms that the separating surface is defined by Eq. (12).

For an orthogonal manipulator such that $a_1 \neq 1$, Eq. (12) is obtained by dividing the DH-parameters by $a_1$. By doing so, we get:

$$a_3 = \frac{1}{2}\sqrt{2a_2{}^2 + 2d_2{}^2 - \frac{2((a_2{}^2 + d_2{}^2)^2 - a_1{}^2(a_2{}^2 - d_2{}^2))}{AB}}$$

where $A = \sqrt{(a_2 + a_1)^2 + d_2{}^2}$ and $B = \sqrt{(a_2 - a_1)^2 + d_2{}^2}$ .

In summary, an orthogonal manipulator given by its four strictly positive DH-parameters $a_1, a_2, a_3, d_2$ has four distinct IKS if and only if,

$$a_3 > \frac{1}{2}\sqrt{2a_2{}^2 + 2d_2{}^2 - \frac{2((a_2{}^2 + d_2{}^2)^2 - a_1{}^2(a_2{}^2 - d_2{}^2))}{\sqrt{(a_2 + a_1)^2 + d_2{}^2}\sqrt{(a_2 - a_1)^2 + d_2{}^2}}} \qquad (14)$$

If $a_1=0$ or $a_2=0$ or $a_3=0$, the manipulator is quaternary [9]. If $d_2=d_3=0$, the manipulator is quaternary if and only if, $a_1 \neq a_2$ and the link lengths do not satisfy $a_1 > a_2 > a_3$ [10].

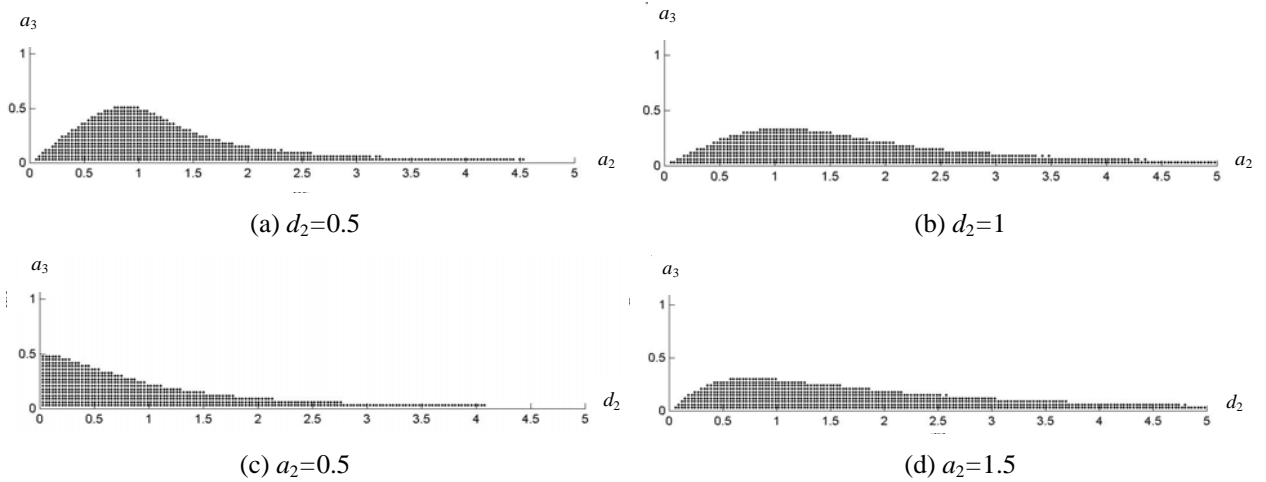

(a) $d_2=0.5$       (b) $d_2=1$

(c) $a_2=0.5$       (d) $a_2=1.5$

**Fig. 6 : Numerical plots of binary manipulators in the same sections as in Fig. 4**

## 6. Conclusion and discussion

A necessary and sufficient condition for an orthogonal manipulator to be quaternary, i.e., to have four distinct inverse kinematic solutions, was established as an explicit expression in the DH-parameters. An orthogonal manipulator given by its four strictly positive DH-parameters $a_1, a_2, a_3, d_2$ is quaternary if and only if,

$$a_3 > \frac{1}{2}\sqrt{2a_2{}^2 + 2d_2{}^2 - \frac{2((a_2{}^2 + d_2{}^2)^2 - a_1{}^2(a_2{}^2 - d_2{}^2))}{\sqrt{(a_2 + a_1)^2 + d_2{}^2}\sqrt{(a_2 - a_1)^2 + d_2{}^2}}} ,$$

and can be assessed easily when designing a new manipulator regional structure. This condition was confirmed numerically by scanning the parameter space. To the authors' knowledge, this condition was never found before.

Figure 7 shows the separating surface in the normalized parameter space ($a_2, a_3, d_2$ are divided by $a_1$). The surface is flat

Wenger P., Chablat D. et Baili M., "A DH-parameter based condition for 3R orthogonal manipulators to have 4 distinct inverse kinematic solutions", Journal of Mechanical Design, Volume 127, pp. 150-155, Janvier 2005.

and close to the plane $a_3$=0, except in the vicinity of $d_2$=0. Binary manipulators, which are below the surface, are much less numerous than their quaternary counterparts.

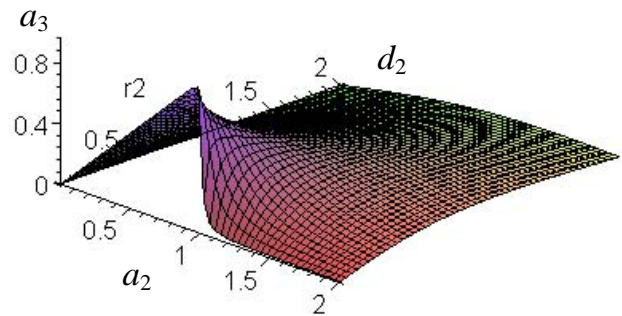

**Fig. 7 : Plot of the separating surface**

This study assumed $d_3$=0, i.e. no offset along the last joint, because our arguments referred to a previous classification of manipulators such that $d_3$=0. But we have noticed in section 4 that the existence condition for a manipulator to have four IKS is independent of $d_3$ and thus Eq. (12) is independent of $d_3$ too. This shows that (12) still plays a role when $d_3 \neq 0$ but this does not prove that condition (14) is still necessary and sufficient. Writing the necessary and sufficient condition for $d_3 \neq 0$ requires to enlarge the classification of [13] and [14], which is under study. We have already found that condition (14) remains necessary and sufficient for any value of $d_3$ provided that $d_2 \geq a_1 / 2\sqrt{2}$, or for $d_3 \leq 2d_2$. On the other hand, it turns out that condition (14) is always sufficient when $d_3 \neq 0$, namely, if (14) is true, then the manipulator will be quaternary for any value of $d_3$.